\documentclass{article}

\PassOptionsToPackage{numbers, compress}{natbib}

\usepackage[preprint]{neurips_2025}

\usepackage[utf8]{inputenc} 
\usepackage[T1]{fontenc}    
\usepackage{hyperref}       
\usepackage{url}            
\usepackage{booktabs}       
\usepackage{amsfonts}       
\usepackage{nicefrac}       
\usepackage{microtype}      
\usepackage{xcolor}         
\usepackage{graphicx}

\usepackage{subcaption}
\usepackage{caption}
\usepackage{wrapfig}
\usepackage{tcolorbox}
\usepackage{amsmath}

\title{Reasoning with \texttt{OmniThought}: A Large CoT Dataset with Verbosity and Cognitive Difficulty Annotations}

\author{
Wenrui Cai$^{1,2}$, Chengyu Wang$^{2*}$, Junbing Yan$^{2}$, Jun Huang$^{2}$, Xiangzhong Fang$^{1}$\\
$^1$ Shanghai Jiao Tong University $^2$ Alibaba Cloud Computing
}

\begin{document}

\maketitle

\begin{abstract}
The emergence of large reasoning models (LRMs) has transformed Natural Language Processing by excelling in complex tasks such as mathematical problem-solving and code generation. These models leverage chain-of-thought (CoT) processes, enabling them to emulate human-like reasoning strategies. However, the advancement of LRMs is hindered by the lack of comprehensive CoT datasets. Current resources often fail to provide extensive reasoning problems with coherent CoT processes distilled from multiple teacher models and do not account for multifaceted properties describing the internal characteristics of CoTs. To address these challenges, we introduce \texttt{OmniThought}, a large-scale dataset featuring \textbf{2 million} CoT processes generated and validated by two powerful LRMs as teacher models. Each CoT process in \texttt{OmniThought} is annotated with novel Reasoning Verbosity (RV) and Cognitive Difficulty (CD) scores, which describe the appropriateness of CoT verbosity and cognitive difficulty level for models to comprehend these reasoning processes. We further establish a self-reliant pipeline to curate this dataset. Extensive experiments using Qwen2.5 models of various sizes demonstrate the positive impact of our proposed scores on LRM training effectiveness. Based on the proposed \texttt{OmniThought} dataset, we further train and release a series of high-performing LRMs, specifically equipped with stronger reasoning abilities and optimal CoT output length and difficulty level. Our contributions significantly enhance the development and training of LRMs for solving complex tasks.
\end{abstract}

\section{Introduction}
In recent years, the field of Natural Language Processing (NLP) has been significantly transformed by the emergence of large language models (LLMs)~\cite{DBLP:journals/corr/abs-2303-18223}, which have demonstrated remarkable proficiency in a wide range of NLP tasks. Of particular interest are large reasoning models (LRMs)~\cite{DBLP:journals/corr/abs-2501-09686}, such as OpenAI's o1\footnote{\url{https://openai.com/o1/}}, DeepSeek-R1~\cite{DBLP:journals/corr/abs-2501-12948}, and QwQ-32B\footnote{\url{https://qwenlm.github.io/blog/qwq-32b/}}, which excel in addressing reasoning challenges such as mathematical problem-solving and code generation through slow thinking processes.

The impressive performance of LRMs can be largely attributed to chain-of-thought (CoT)~\cite{DBLP:conf/nips/Wei0SBIXCLZ22}. CoT empowers LRMs to break down intricate problems into intermediate steps, closely emulating human problem-solving~\cite{DBLP:conf/emnlp/HaoGMHWWH23,DBLP:conf/emnlp/Yan0ZHHZ23}. This capability supports various applications, including science education~\cite{DBLP:conf/aaai/CohnHLB24}, robotic control~\cite{DBLP:conf/corl/ZawalskiCPMFL24}, and clinical assessment~\cite{DBLP:journals/artmed/GuJPY25}, among others. Producing effective LRMs necessitates CoT-based training, which typically incorporates a synergy of supervised and reinforcement learning. Through supervised fine-tuning (SFT), models develop the ability to follow explicit reasoning patterns, where both the CoT process and solutions are treated as outputs, while reinforcement learning (RL) enables them to optimize reasoning strategies through iterative feedback and exploration~\cite{DBLP:journals/corr/abs-2410-01679,DBLP:journals/corr/SchulmanWDRK17,DBLP:journals/corr/abs-2402-03300,DBLP:conf/acl/TrungZJSJL24}.

\begin{figure}[t]
\centering
\includegraphics[width=0.85\linewidth]{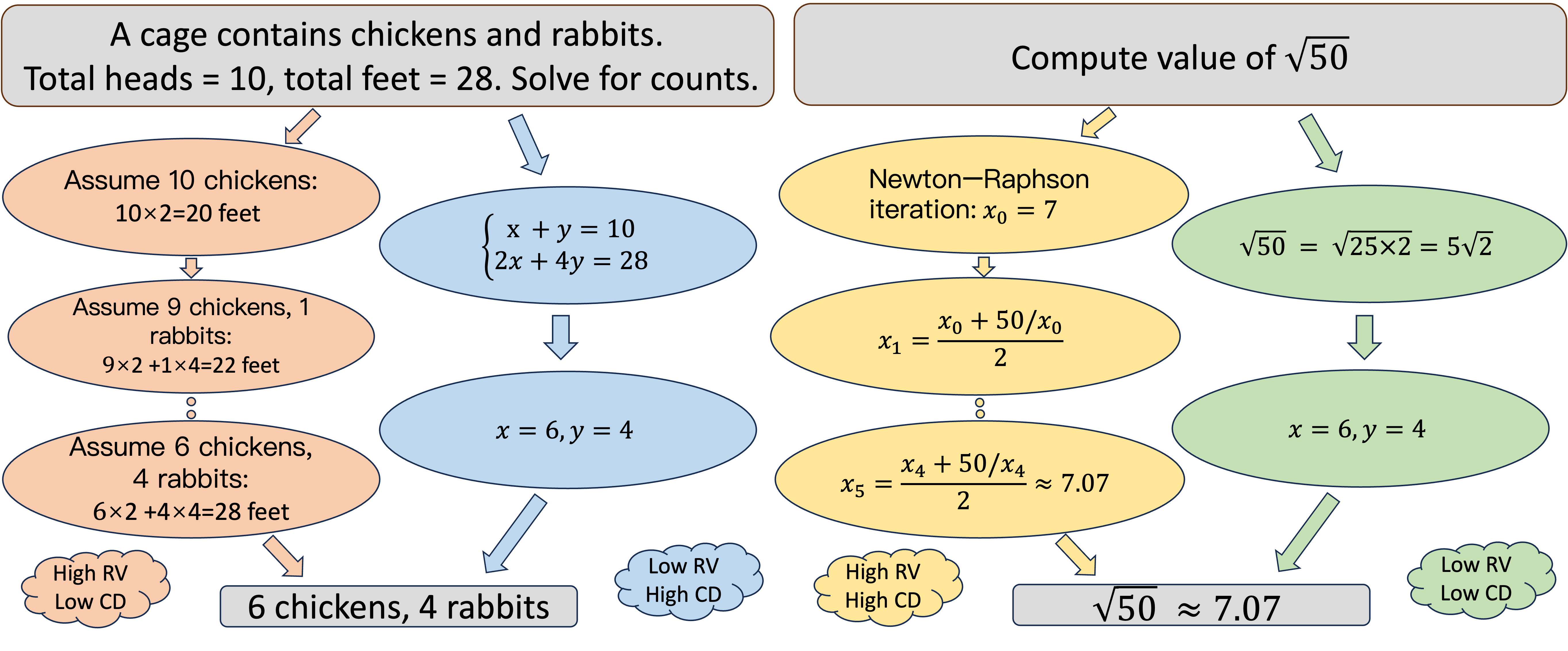}
\caption{A motivation example of CoTs with different Reasoning Verbosity (RV) and Cognitive Difficulty (CD) levels. For simplicity, we only present key steps in these CoTs.}
\vspace{-1em}
\label{fig:cot}
\end{figure}

Despite these advances, the development of LRMs is impeded by the lack of large-scale comprehensive CoT datasets. Current data resources within the open-source community, often directly distilled from powerful LRMs, fall short of providing a sufficient number of reasoning problems with detailed CoT processes generated by multiple teachers and multifaceted properties describing the characteristics of these CoTs. Numbers of CoT processes in these datasets are compared in Table~\ref{tab:num}. Beyond the dataset size, we propose the following aspects that should be specifically emphasized:

\begin{wraptable}{r}{0.475\textwidth}
    \centering
    \caption{A comparison of the numbers of CoT processes in popular open-source datasets (see Hugging Face Datasets) for reasoning-related problems (such as mathematics, code).}
    \begin{small}
    \begin{tabular}{l|c}
        \hline
        \textbf{Dataset Name} & \textbf{\#CoTs (K)} \\
        \hline
        OpenThoughts-114k & 114 \\
        GeneralThought-430K & 430 \\
        SYNTHETIC-1-SFT-Data & 894 \\
        OpenThoughts2-1M & 1,143 \\
        AM-DeepSeek-R1-Distilled-1.4M & 1,400 \\
        \hline
        \bf\texttt{OmniThought} (Ours) & 2,059 \\
        \hline
    \end{tabular}
    \end{small}
    \label{tab:num}
\end{wraptable}

\textbf{Comprehensive Quality Assessment.} Based on previous research~\cite{DBLP:conf/acl/JacoviBBHHT0AG24}, logical errors in CoT-based training sets negatively impact the performance of resultant LRMs. Examining the logical correctness of CoT processes within the ``LLM-as-a-judge'' paradigm~\cite{DBLP:journals/corr/abs-2411-15594} is essential, as human annotations are impractical. Validating the correctness of CoT processes with LLM judges is vital to ensure the quality of the datasets. Furthermore, collecting and releasing CoT processes of lower quality can also be valuable to support other training algorithms for LRMs, such as direct preference optimization (DPO)~\cite{DBLP:conf/nips/RafailovSMMEF23} and reward model training~\cite{DBLP:conf/nips/Ouyang0JAWMZASR22}. To our knowledge, these features are not fully explored in existing dataset-related works.

\textbf{Reasoning Verbosity (RV).} It is evident that there are no ground-truth CoTs for reasoning problems, with numerous candidates that can be marked as correct or feasible model outputs. Yet, using excessively long CoTs as training data can impair the reasoning performance of the resulting models~\cite{DBLP:journals/corr/abs-2502-18080,DBLP:journals/corr/abs-2501-12570}. Therefore, a series of studies have discovered that there are optimal CoT lengths for different types of reasoning problems~\cite{DBLP:journals/corr/abs-2502-18080}. In our work, we propose and compute the Reasoning Verbosity (RV) score for each problem-CoT pair, which guides users in selecting an appropriate subset of CoTs to train LRMs with length-optimal CoTs. This improves model performance while avoiding excessive reasoning, without the need for costly test-time inference scaling.

\textbf{Cognitive Difficulty (CD).} As discussed, the RV score of a CoT is measured considering the complexity of the problem, regardless of the target models to be trained. Since our dataset is constructed for training LRMs of different sizes, the CoTs should also be measured based on the capacities of the underlying LRMs. We suggest that the cognitive difficulty of a CoT affects its suitability for LRMs of various sizes. For example, smaller models often exhibit distinct capabilities and cognitive trajectories compared to their larger counterparts when solving reasoning tasks~\cite{DBLP:conf/emnlp/Yan0ZHHZ23,DBLP:conf/iclr/0006LCF24,DBLP:journals/corr/abs-2502-18001}. Thus, overly complicated reasoning steps (which are not necessarily difficult), though logically correct, may not be suitable for the learning procedures of smaller models to acquire the necessary abilities. We propose that cognitive difficulty can guide more effective LRM training based on selected backbones of varied parameter sizes. For example, utilizing relatively simpler CoTs is more suitable for training a small model, as they are more suitable for the small model’s cognitive ability and more economical for inference. An example of CoTs with different RV and CD levels for the same reasoning problem is shown in Figure~\ref{fig:cot}.

\begin{figure}[t]
\centering
\includegraphics[width=.925\linewidth]{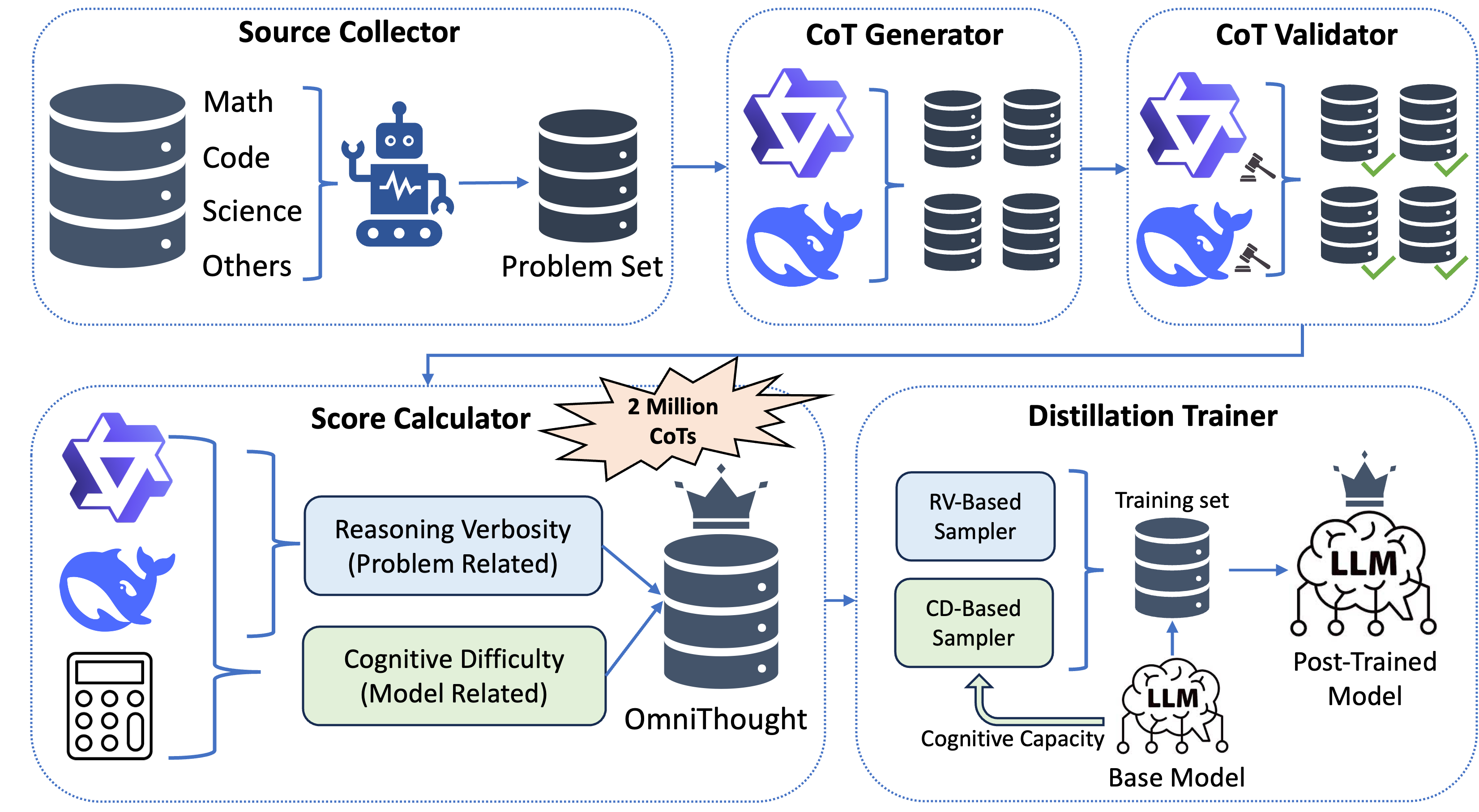}
\caption{Our framework, including the dataset construction pipeline and the training procedure. Note: We leverage {DeepSeek-R1} and {QwQ-32B} as the teacher models in all modules of our work. Other sufficiently strong models can also be leveraged as teacher models.}
\vspace{-1em}
\label{fig:framework}
\end{figure}

To address these limitations, we introduce \texttt{OmniThought}\footnote{\url{https://huggingface.co/datasets/alibaba-pai/OmniThought}}, which comprises challenging problems with \textbf{2 million} CoT processes generated and validated by powerful LRMs and annotated with the novel RV and CD scores. To create the dataset, we propose a self-reliant pipeline to curate the \texttt{OmniThought} dataset, as shown in Figure~\ref{fig:framework}. First, the \texttt{Source Collector} aggregates reasoning problems from established open datasets, ensuring a broad and varied collection of tasks. Next, the \texttt{CoT Generator} employs multiple LRMs to generate CoT processes for each task, with the \texttt{CoT Validator} verifying the logical structure and correctness of each CoT process. Finally, the \texttt{Score Calculator} generates RV and CD scores, indicating the characteristics of CoTs.

In our experiments, we validate the effectiveness of \texttt{OmniThought} through extensive benchmarking. We utilize Qwen2.5 models~\cite{DBLP:journals/corr/abs-2412-15115} with varying parameter sizes as our backbones. The results demonstrate that the proposed RV and CD scores positively impact the efficacy of LRM training. Based on the CoT sampler for \texttt{OmniThought}, we further train and release a series of strong LRMs (including 7B and 32B versions). These models excel in reasoning abilities with optimal CoT length and difficulty level, even outperforming the DeepSeek-R1-Distill series trained using proprietary datasets~\cite{DBLP:journals/corr/abs-2501-12948}. In summary, we make the following contributions:
\begin{itemize}
    \item From the~\textbf{data} perspective, we construct~\texttt{OmniThought}, a large-scale dataset composed of reasoning problems with \textbf{2 million} CoTs sourced and validated from multiple teachers. The CoTs are annotated with RV and CD scores, reflecting the characteristics of CoTs.
    
    \item From the~\textbf{method} perspective, we propose a self-reliant pipeline to curate the \texttt{OmniThought} dataset without manual intervention. Based on the dataset, we introduce a series of practical guidelines for CoT samplers to select optimal CoTs for training powerful LRMs. 
    
    \item From the~\textbf{model} perspective, we validate the effectiveness of the \texttt{OmniThought} dataset through extensive experiments. We further release a series of LRMs based on \texttt{OmniThought}, which excel in reasoning abilities with optimal CoT length and difficulty level.
\end{itemize}


\section{Related Work}

Generating CoT processes has garnered significant attention in recent years due to its potential to enhance the reasoning capabilities of LLMs. Early work often relied on manual annotation performed by domain experts to produce gold-standard CoTs for benchmarking~\cite{DBLP:conf/nips/HuangWXLZXFYCY024,DBLP:journals/corr/abs-2410-07985}. These methods ensure high-quality outputs but are labor-intensive and limited in scalability. Automatic approaches have focused on leveraging LLMs through prompt engineering (e.g., \emph{Let's think step by step}), tapping into the models' inherent capabilities to generate detailed reasoning processes~\citep{DBLP:conf/nips/Wei0SBIXCLZ22,DBLP:conf/nips/YaoYZS00N23,DBLP:conf/emnlp/LiuG0HZQZ23,DBLP:conf/iclr/0002WSLCNCZ23}. While these techniques can rapidly produce CoT data, they are often constrained by model biases and require careful prompt design. The advent of powerful LRMs, such as DeepSeek-R1~\cite{DBLP:journals/corr/abs-2501-12948}, has led to new approaches that directly generate CoT processes~\cite{cai2025trainingsmallreasoningllms,openthoughts}. Despite these advancements, significant challenges remain in generating high-quality CoT processes, particularly in addressing varying difficulty levels and aligning with model capabilities.

Recently, several works have discovered that generated CoT processes may not be optimal in various aspects. For example, \cite{DBLP:journals/corr/abs-2502-18080} shows that LRMs sometimes generate overly lengthy CoTs, which can harm reasoning performance. \cite{DBLP:journals/corr/abs-2502-18001} demonstrates that stronger models gain advantages from detailed reasoning, whereas less powerful models exhibit improved performance with simple CoT supervision. Several projects have begun exploring concise yet effective CoTs, such as those implemented in \texttt{DeepSeek-V3-0324}\footnote{\url{https://huggingface.co/deepseek-ai/DeepSeek-V3-0324}}. Subsequent research has further investigated switchable reasoning modes, where models can toggle between normal output and CoT reasoning, such as Llama-Nemotron~\cite{bercovich2025llamanemotron} and Qwen3\footnote{\url{https://qwenlm.github.io/blog/qwen3/}}. In contrast, our work suggests that lengthy and concise CoTs represent different reasoning forms, and models should automatically choose the optimal CoT based on task complexity and its cognitive ability. Challenging problems may require deeper reasoning, while simpler ones might not.


\section{Construction of the~\texttt{OmniThought} Dataset}

In this section, we introduce the construction process of the proposed \texttt{OmniThought} dataset (as shown in Figure~\ref{fig:framework}). The pipeline consists of the following components: the \texttt{Source Collector}, the \texttt{CoT Generator}, the \texttt{CoT Validator}, and the \texttt{Score Calculator}. The \texttt{Source Collector} is responsible for gathering reasoning problems. The \texttt{CoT Generator} and the \texttt{CoT Validator} are used to generate and validate CoT processes using multiple teacher models. Finally, the \texttt{Score Calculator} computes Reasoning Verbosity (RV) and Cognitive Difficulty (CD) scores.


\subsection{Basic Modules: \texttt{Source Collector}, \texttt{CoT Generator} and \texttt{CoT Validator}}

\begin{wrapfigure}{r}{0.35\textwidth}
    \centering
    \includegraphics[width=\linewidth]{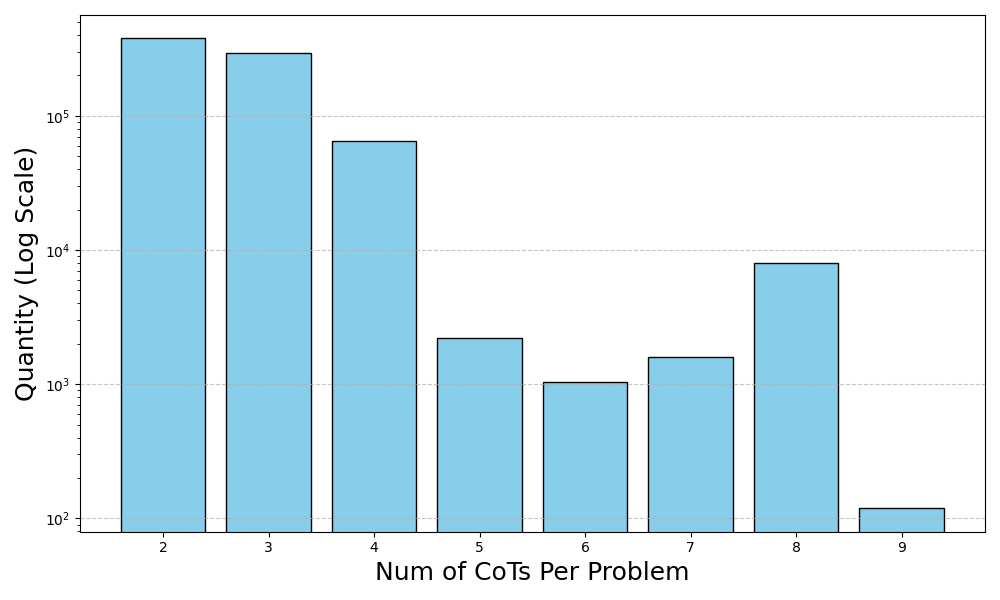}
    \caption{The distribution of CoT processes per problem.}
    \label{fig:dist}
\end{wrapfigure}

The \texttt{Source Collector} firstly selects a diverse range of reasoning problem sources. It is worth mentioning that many high-quality reasoning problem sets are available within the open-source community. For this study, we leverage OpenThoughts2-1M\footnote{\url{https://huggingface.co/datasets/open-thoughts/OpenThoughts2-1M}} and DeepMath-103K~\cite{he2025deepmath103k}. OpenThoughts2-1M contains approximately 640K reasoning problems across various domains, such as mathematics, coding, science, and puzzles, while DeepMath-103K comprises 103K mathematical problems of varying difficulty levels. We combine these two datasets to form the reasoning problem set for the \texttt{OmniThought} dataset.

Next, the \texttt{CoT Generator} utilizes DeepSeek-R1 and QwQ-32B as teacher models to generate multiple reasoning processes for the problem set provided by the \texttt{Source Collector}. To ensure the high quality of the generated CoT processes, the \texttt{CoT Validator} employs the ``LLM-as-a-judge'' approach to validate the correctness of the generated CoT processes from multiple aspects, including logical correctness and the ability to derive the correct answer. The prompt template is shown in Appendix~\ref{sec:prompt}.

As mentioned earlier, releasing CoT processes of lower quality can also be valuable, as they can support DPO algorithms~\cite{DBLP:conf/nips/RafailovSMMEF23} or reward model training~\cite{DBLP:conf/nips/Ouyang0JAWMZASR22}. Therefore, we retain the process in the dataset unless it produces incorrect answers. It is important to note that the OpenThoughts2-1M and DeepMath-103K datasets already include CoT processes from DeepSeek-R1, and we use the \texttt{CoT Validator} to re-validate these processes, where the \texttt{CoT Validator} employs the same language reasoning models (LRMs) as the \texttt{CoT Generator}. We add the validation results as metadata.

Ultimately, the \texttt{OmniThought} dataset contains more than \textbf{2 million} CoT processes for 708K reasoning problems. We ensure that each problem in the dataset has at least two validated correct CoT processes. The distribution of CoT processes per problem in our dataset is shown in Figure~\ref{fig:dist}. The sample data format is presented in Appendix~\ref{sec:sample}.

\subsection{Reasoning Verbosity: Definition, Calculation and Verification}

The \texttt{Score Calculator} evaluates the RV and CD scores for CoT processes, which is the core module of our pipeline. We begin with an introduction to the RV score.

CoT processes naturally involve self-reflection, prompting models to undergo multiple rounds of reflection and correction during reasoning~\cite{DBLP:journals/corr/abs-2310-06692}. This mechanism reduces errors in complex problems but can lead to ``overthinking'' in simpler ones~\cite{DBLP:journals/corr/abs-2501-18585}, such as excessive checks on ``1 + 1 = ?''. This overthinking wastes computational resources and decreases reasoning accuracy. Specifically, for a given problem, the length of its CoT should align with the problem's difficulty, reflecting the ``Reasoning Verbosity'' (RV) of the CoT. This phenomenon has also been observed in concurrent work~\cite{DBLP:journals/corr/abs-2502-18080}. However, it remains unclear whether and how RV impacts the training effectiveness of LRMs. We formally define RV grading criteria on a 0 to 9 scale:

\begin{footnotesize}
\begin{tcolorbox}[colback=blue!5!white, colframe=blue!60!black, title=Grading Criteria for Reasoning Verbosity]
0-1: Minimal verbosity, straightforward expression with little to no elaboration.

2-3: Clear and concise reasoning with necessary explanations.

4-5: Moderate verbosity with detailed explanations and thorough reasoning.

6-7: Extensive verbosity with comprehensive justification and exploration of complex connections.

8-9: High verbosity with deep, exhaustive exploration of reasoning; involves extensive elaboration, nested justifications, and consideration of counterarguments or alternative perspectives.
\end{tcolorbox}
\end{footnotesize}

\begin{wrapfigure}{r}{0.285\textwidth}
    \centering
    \begin{subfigure}[b]{0.285\textwidth}
        \centering
        \includegraphics[width=\textwidth]{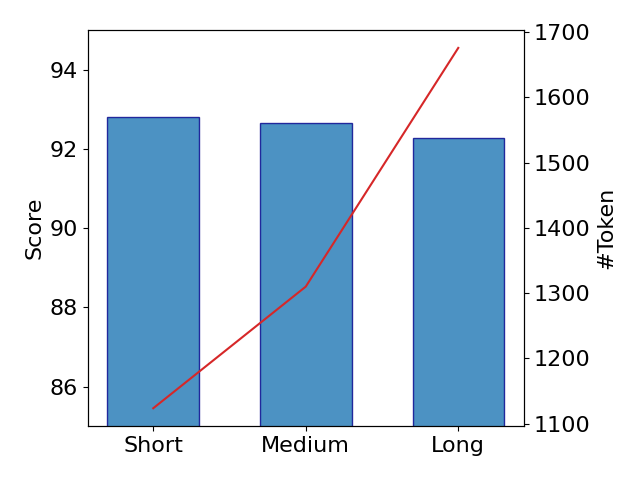}
        \caption{GSM8K}
        \label{fig:sub1}
    \end{subfigure}
    \hfill
    \begin{subfigure}[b]{0.285\textwidth}
        \centering
        \includegraphics[width=\textwidth]{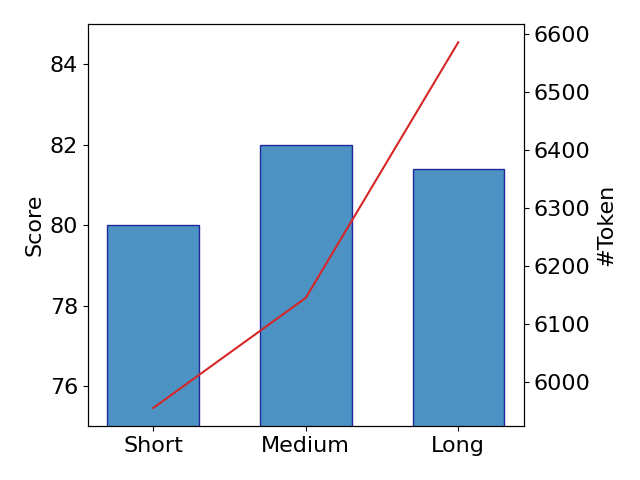}
        \caption{MATH500}
        \label{fig:sub2}
    \end{subfigure}
    \hfill
    \begin{subfigure}[b]{0.285\textwidth}
        \centering
        \includegraphics[width=\textwidth]{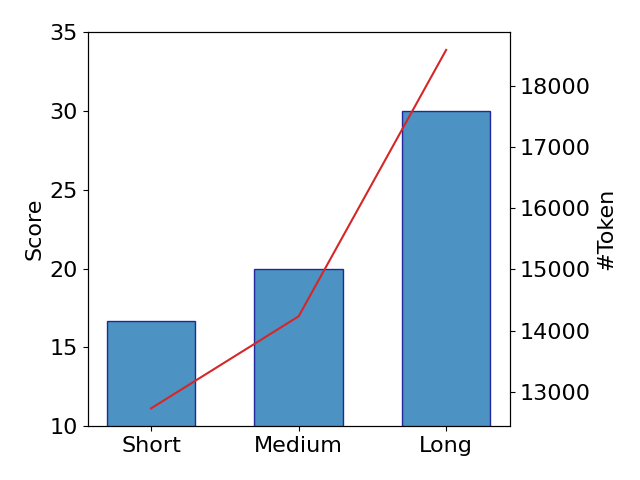}
        \caption{AIME24}
        \label{fig:sub3}
    \end{subfigure}
    \caption{Comparison using CoTs from varying RV levels as training sets.}
    \label{fig:rv_exp}
\end{wrapfigure}

We observe that, in addition to the ``LLM-as-a-judge'' paradigm, the number of output tokens in the CoT process itself provides sufficient hints for assessing verbosity. However, these two judgment systems do not always yield consistent evaluations. For instance, a CoT process might be verbose in terms of steps (having a large number of reasoning steps), but if each step is relatively simple, the total number of output tokens may not necessarily be large. Let $L$ be the token count of the CoT process. We also normalize $L$ to a scale of 0 to 9 (denoted as $L_{\text{norm}}$) as follows:
\begin{equation}
    L_{\text{norm}} = \mathcal{K} \cdot \frac{\log(L - L_{\text{min}} + 1)}{\log(L_{\text{max}} - L_{\text{min}} + 1)}
\end{equation}
where $L_{\text{min}}$ and $L_{\text{max}}$ represent the minimum and maximum token counts of CoT processes in our dataset, and $\mathcal{K}$ is the grading scale (i.e., $\mathcal{K}=9$ in our case). Next, the knowledge of $L_{\text{norm}}$ and the vanilla RV score judged by LLMs are fused into the final RV score $S_{RV}$, defined as: $S_{RV} = \text{round}(\alpha \cdot L_{RV} + (1 - \alpha) \cdot L_{\text{norm}})$
where $\alpha \in (0,1)$ reflects the relative importance of the two features (empirically set to 0.5), and $L_{RV}$ is the model judgment score (please refer to the prompt template in Appendix~\ref{sec:prompt}). CoT examples of different RV levels are shown in Appendix~\ref{sec:cotsample}.

\textbf{Experimental Verification.} To further verify the effectiveness of RV in training LRMs, we randomly sample a subset consisting of 10K problems, each having three CoT processes categorized into three distinct RV levels. In this subset, the RV difference between adjacent levels exceeds 3. Based on this arrangement, we construct three training datasets with the same problems but different RV scores. Next, we execute SFT training initialized from Qwen2.5-7B-Instruct on each dataset with identical configurations (see Appendix~\ref{sec:exp}) to produce three models: short, medium, and long.

We evaluate these models across GSM8K~\cite{DBLP:journals/corr/abs-2110-14168}, MATH500~\cite{DBLP:conf/nips/HendrycksBKABTS21}, and AIME24\footnote{\url{https://artofproblemsolving.com/wiki/index.php/2024_AIME_I}}. From the results illustrated in Figure~\ref{fig:rv_exp}, the models effectively assimilate the characteristics embedded within the SFT training set. Moreover, on the relatively simple GSM8K tasks, all models demonstrate similar performance; an increase in output token count does not enhance accuracy and even results in a slight decline. On the medium-difficulty MATH500 tasks, accuracy initially increases with the token count and subsequently declines, with the medium-model attaining the highest accuracy while generating a moderate token count. In the most challenging AIME24 problems, the long-model achieves the highest score; model accuracy increases with token count and exhibits significant improvements once the token count surpasses a certain threshold.

These experimental results confirm our hypothesis: for difficult problems, a longer CoT process can correct the model’s own errors, thereby effectively improving accuracy. However, in simple tasks, excessive reasoning and verification in CoTs not only increase computational resource consumption but may also reduce problem-solving accuracy. Therefore, the verbosity of the CoT process should be matched to the problem’s difficulty, which is the starting point for our proposal of the RV score. Ultimately, we can construct training sets with CoT processes at appropriate RV levels according to task difficulty, thereby maximizing computational resource utilization while ensuring high accuracy.

\subsection{Cognitive Difficulty: Definition, Calculation and Verification}

\begin{wraptable}{r}{0.4\textwidth}
    \centering
    \caption{Comparison of CoT difficulty on MATH500 in terms of model scales.}
    \begin{small}
    \begin{tabular}{l|c}
        \hline
        \textbf{Model} & \textbf{Avg. Score} \\
        \hline
        DS-R1-Distill-Qwen-1.5B & 4.5 \\
        DS-R1-Distill-Qwen-7B & 6.2 \\
        DS-R1-Distill-Qwen-32B & 7.3 \\
        \hline
    \end{tabular}
    \end{small}
    \label{tab:cd_exp}
\end{wraptable}

Furthermore, we argue that in the dataset, the difficulty of CoTs should align with the cognitive abilities of the target model to be trained. Due to significant differences in model parameter sizes, the cognitive and reasoning trajectories of large and small models are not always identical~\cite{DBLP:conf/emnlp/Yan0ZHHZ23,DBLP:conf/iclr/0006LCF24}. Smaller models, constrained by their parameter limits, tend to rely on more basic methods to solve problems, while larger models, with more advanced cognitive abilities, may employ higher-level techniques. For instance, For the problem of calculating the area of a triangle from given coordinates, a small model may employ simple formulaic geometric decomposition, while a larger model may apply more sophisticated methods, such as vector-based algebraic abstraction. 


\textbf{Experimental Verification.}
To test the hypothesis, we conduct an experiment using three models from the DeepSeek-R1-Distill series~\cite{DBLP:journals/corr/abs-2501-12948}: DeepSeek-R1-Distill-Qwen-1.5B, DeepSeek-R1-Distill-Qwen-7B, and DeepSeek-R1-Distill-Qwen-32B. We evaluate these models on MATH500~\cite{DBLP:conf/nips/HendrycksBKABTS21}. For each model's CoT processes, we employ QwQ-32B to assign a difficulty rating from 0 to 9 based on the methodological complexity and overall intricacy of reasoning (refer to our criteria below). Each CoT process is evaluated three times, and the scores are averaged to produce the final rating. We then average the difficulty ratings for each model to obtain its overall difficulty score, presented in Table~\ref{tab:cd_exp}.

\begin{wrapfigure}{r}{0.375\textwidth}
    \centering
    \includegraphics[width=\linewidth]{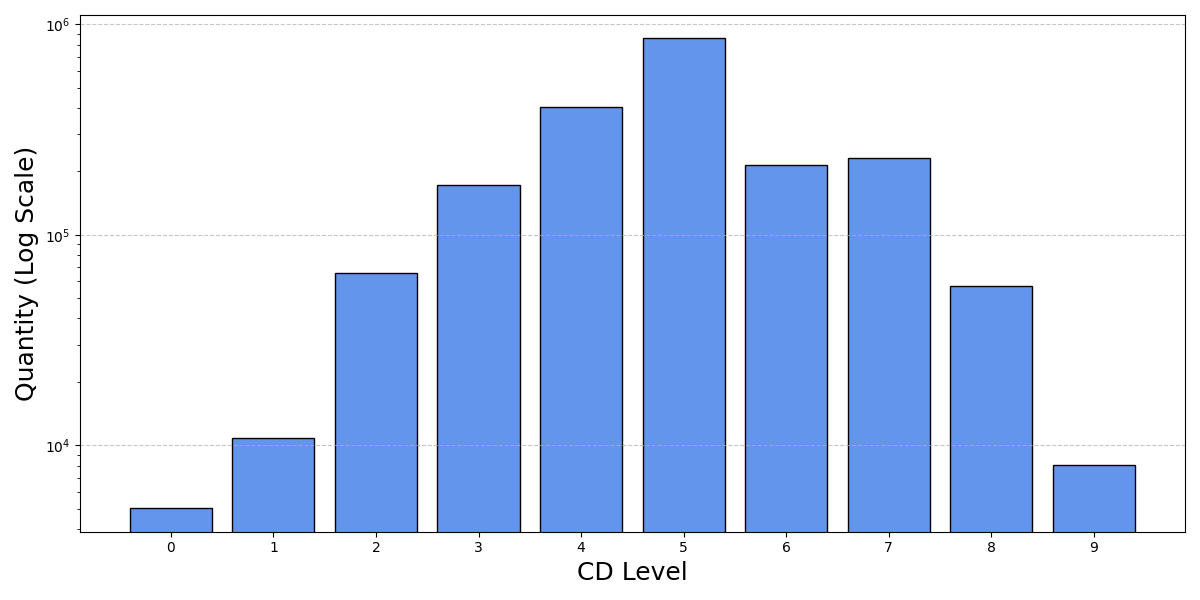}
    \caption{The distributions of CD scores of CoTs in \texttt{OmniThought}.}
    \label{fig:cd_dist}
\end{wrapfigure}

The results demonstrate that the difficulty of CoT processes escalates as the model size increases, suggesting that larger models possess enhanced reasoning and cognitive capacities. Consequently, difficult CoT processes may not be suitable for training models with lower cognitive abilities. It is therefore essential to use CoT processes that align with the model's cognitive trajectory to improve its reasoning capabilities, a strategy akin to ``teaching according to the student's ability.'' In our work, the CD score considers the difficulty of the methods used in the CoT. The grading criteria, on a scale from 0 to 9, are shown below. The prompt template, along with CoT examples for each CD level, is presented in Appendix~\ref{sec:prompt} and Appendix~\ref{sec:cotsample}, respectively.

\begin{footnotesize}
\begin{tcolorbox}[colback=green!5!white, colframe=green!50!black, title=Grading Criteria for Cognitive Difficulty]
0-1: Elementary facts or a single trivial operation.

2-3: Multi-step arithmetic, explicit enumeration, basic rule chaining.

4-5: Early-undergraduate logic/algebra; one non-obvious insight.

6-7: Advanced undergraduate techniques (determinants, dynamic programming, code reasoning, etc.).

8-9: Graduate-level abstraction, nested proofs, intricate algorithmic analysis.
\end{tcolorbox}
\end{footnotesize}

In \texttt{OmniThought}, we score all the validated correct CoT processes. The CD distributions are shown in Figure~\ref{fig:cd_dist}. It can be observed that the distribution follows a Gaussian-like pattern, peaking at levels 4–5 and gradually decreasing toward both extremes. This finding also indicates that highly capable models, such as DeepSeek-R1 or QwQ-32B, have a probability of producing extremely difficult CoT processes. When performing knowledge distillation, models with limited cognitive capacity are unlikely to effectively comprehend these processes. Therefore, given an original CoT dataset and a base model, one can filter the training dataset according to the base model’s cognitive capability, thereby maximizing the utilization of existing CoT data and effectively improving the model’s reasoning ability.

\subsection{Further Explorations on Reasoning Verbosity and Cognitive Difficulty}

Based on the above RV and CD scores, we further propose the following two research questions.

\underline{\textbf{RQ1:} RV vs. CD, which is more significant for LRM training, and is a combined approach superior?}

To investigate this question, we conduct the following study. First, we randomly sample a subset of 10K problems from \texttt{OmniThought}, where each problem has at least four CoT processes, and the maximum differences in RV and CD among these processes both exceed 4. We believe that under this configuration, the CoT processes for each problem exhibit significant differences in both verbosity and difficulty, making them suitable as experimental subjects. Given an initial LLM, we employ four different strategies to select the SFT training set: optimal RV, optimal CD, the combined approach, and random selection. In the RV-optimal method, we define an RV range and select the CoT processes that best fit this range for inclusion in the training set. In the CD-optimal method, we define a CD range and construct the dataset accordingly. In the combined approach, we specify both RV and CD ranges and weight RV and CD conformity equally to select the CoT processes. In random selection, we randomly choose one CoT process per problem to build the training set.

We employ Qwen2.5-7B-Instruct as the base model, with an RV range of 3–5 and a CD range of 0–6, which are empirically suitable for the 7B model to learn. After constructing the four datasets, we train four models with identical SFT configurations (see Appendix~\ref{sec:exp}). We then evaluate these models on AIME24, MATH500, GPQA-Diamond~\cite{DBLP:journals/corr/abs-2311-12022}, and LiveCodeBench V2~\cite{DBLP:conf/iclr/JainHGLYZWSSS25}; the results are shown in Table~\ref{tab:select}. Compared to random selection, both RV-based and CD-based selections yielded greater improvements, with the combined selection achieving the highest scores. Among RV and CD scores, CD plays a more important role. By adjusting the RV and CD ranges accordingly, we can obtain training sets that align more closely with the model’s inherent cognitive ability.

\begin{table}[t]
\centering
\caption{The impact of different CoT selection strategies on reasoning performance.}
\begin{small}
\begin{tabular}{l|llll|l}
\hline
\bf Strategy & \bf AIME2024 & \bf MATH500 & \bf GPQA Diamond & \bf LiveCodeBench V2 & \bf Average\\
\hline
Random & 16.67 & 80.6 & 36.36 & 31.31 & 41.24\\
\hline
RV Optimal  & 26.67 & 83.2 & \underline{40.40} &  34.44 & 46.18\\
CD Optimal  & \underline{33.33} & \underline{83.8} & 39.90 & \underline{36.10} & \underline{48.28}\\
Combined & \bf 36.67 & \bf 84.4 & \bf 40.91 & \bf 36.59 & \bf 64.25\\
\hline
\end{tabular}
\end{small}
\label{tab:select}
\vspace{-1em}
\end{table}

\underline{\textbf{RQ2:} how does our dataset benefit other training algorithms?}

\begin{wraptable}{r}{0.5\textwidth}
    \centering
    \caption{Performance comparison before and after DPO training. The numbers in parentheses indicate the numbers of average output tokens per problem. }
    \begin{small}
    \begin{tabular}{l|c|c}
        \hline
        \textbf{Dataset} & \textbf{Model w/o. DPO} & \textbf{Model w. DPO}\\
        \hline
        AIME2024 & 36.67 (12,248) & 36.67 (10,352)\\
        MATH500	& 84.4 (3,676) & 86.2 (3,108)\\
        GPQA-D & 40.91 (6,295) & 42.93 (5,635)\\
        LCB V2 & 36.59 (8,599) & 39.9 (7,658)\\
        \hline
    \end{tabular}
    \end{small}
    \label{tab:dpo}
\end{wraptable}

Beyond the commonly used SFT, \texttt{OmniThought} can also be effectively applied to other training algorithms. Taking the DPO algorithm~\cite{DBLP:conf/nips/RafailovSMMEF23} as an example, criteria can be defined for chosen and rejected CoT processes to construct a preference-pair dataset. Within the 10K problem subset constructed for RQ1, we regard CoT processes with an RV range of 3–5 as chosen and those with the maximum RV as rejected, thereby constructing 10K preference pairs. Starting from the SFT model from RQ1, we apply DPO algorithm. We evaluate the resulting model on AIME24 and MATH500 and record its output token counts; the results are presented in Table~\ref{tab:dpo}. It can be observed that: (i) the model’s score shows further improvement on MATH500, GPQA-D and LiveCodeBench V2; (ii) the results on AIME24 remain unchanged; and (iii) the output token counts of the new model decrease on all benchmarks. Therefore, by selecting CoT processes with desirable and undesirable scores as chosen and rejected responses, we can further adjust the model's output preferences without harming its accuracy.






\section{Training Strong LRMs on the~\texttt{OmniThought} Dataset}

In this section, we introduce our procedures to develop a series of LRMs based on \texttt{OmniThought}, which are equipped with strong reasoning abilities with optimal CoT length and difficulty level.

\subsection{CoT Selection Guidelines}

Based on our goal, the selected CoTs from the entire dataset satisfy two criteria: (i) the RV score should be as optimal as possible, and (ii) the CD score should align with the student model's capacity. Denote $\mathcal{D} = \{(x, y_{CoT}, y, S_{RV}, S_{CD})\}$ as the proposed dataset, where $x$, $y_{CoT}$, $y$, $S_{RV}$, and $S_{CD}$ represent the problem, the CoT process, the answer, and the RV and CD scores, respectively. For any problem $x$, a collection of CoTs with their metadata can be obtained, denoted as $\mathcal{D}(x)$. We further denote the model capacity score as $\mu_{CD}$\footnote{Please refer to the experiments below as an example to determine the value of $\mu_{CD}$.}. We select the most suitable CoT with the probability:
\begin{equation}
P_1(y_{CoT}) \propto 
\begin{cases}
\max_{i=1,\ldots,|\mathcal{D}(x)|}\bigl\lvert S^{(i)}_{CD}-\mu_{CD}\bigr\rvert, & S_{CD}\le\mu_{CD},\\[0.5em]
\max_{i=1,\ldots,|\mathcal{D}(x)|}\bigl\lvert S^{(i)}_{CD}-\mu_{CD}\bigr\rvert - \bigl(S_{CD}-\mu_{CD}\bigr), & S_{CD}>\mu_{CD}.
\end{cases}
\end{equation}
This implies that CoTs with $S_{CD}\le\mu_{CD}$ are assigned an equally high probability of being selected, whereas for those with $S_{CD}>\mu_{CD}$, the greater the deviation from $\mu_{CD}$, the lower the probability. We aim to select \(y_{CoT}\) that falls within the model’s cognitive capacity ($\mu_{CD}$) with a higher probability.

In addition, the CD and RV scores are naturally related in that difficult solutions typically require verbose explanations. Therefore, we need to penalize situations where, for a specific CoT, the score gap between CD and RV is excessively high. Empirically, we define the second rule as follows:
\begin{equation}
    P_2(y_{CoT}) \propto \left(\max_{i=1,\ldots,\vert\mathcal{D}(x)\vert}\vert S^{(i)}_{CD}-S^{(i)}_{RV}\vert\right) - \vert S_{CD}-S_{RV}\vert.
\end{equation}
Based on these two rules, suitable CoTs can be sampled to form training sets. We refer readers to the detailed sampling process in Appendix~\ref{sec:sample}.

\begin{table}[t]
\centering
\caption{Performance comparison between our trained models and state-of-the-art distilled LRMs in the open-source community. The best scores are printed in \textbf{bold} with the second best \underline{underlined}.}
\begin{small}
\begin{tabular}{l|llll|l}
\hline
\bf Model & \bf AIME2024 & \bf MATH500 & \bf GPQA-D & \bf LCB V2 & \bf Avg.\\
\hline
OpenThinker-7B & 31.3 & 83.0 & 42.4 & 39.9 & 49.1\\
DeepSeek-R1-Distill-Qwen-7B~\cite{DBLP:journals/corr/abs-2501-12948} & \bf 57.3 & \underline{89.6} & 47.3 & 48.4 & 60.6\\
OpenThinker2-7B  & 50.0 & 88.4 & \underline{49.3} & \underline{55.6} & \underline{60.8}\\
\bf \texttt{OmniThoughts-7B} (Ours) & \underline{56.7} & \bf 90.2 & \bf 50.0 & \bf 56.8 
 & \bf 63.4\\
\hline
LIMO-32B~\cite{DBLP:journals/corr/abs-2502-03387} & 56.7 & 86.6 & 58.1 & 60.0 & 65.3\\
OpenThinker-32B & 66.0 & 90.6 & 61.6 & 68.9 & 71.7\\
DeepSeek-R1-Distill-Qwen-32B~\cite{DBLP:journals/corr/abs-2501-12948} & 74.7 & 90.0 & 62.4 & 72.3 & 74.8\\
OpenThinker2-32B  & \underline{76.7} & \underline{90.8} & \bf 64.1 & \underline{72.5} & \underline{76.0}\\
Light-R1-32B~\cite{DBLP:journals/corr/abs-2503-10460} & 74.7 & 90.4 & 62.0 & 56.0 & 70.7\\
s1.1-32B~\cite{DBLP:journals/corr/abs-2501-19393} & 59.3 & 87.4 & 62.0 & 58.7 & 66.8\\
\bf \texttt{OmniThoughts-32B} (Ours) & \bf 80.0 & \bf 92.6 & \underline{64.0} & \bf 73.4 & \bf 77.5\\
\hline
\end{tabular}
\end{small}
\label{tab:strong}
\vspace{-.5em}
\end{table}

\subsection{Major Experimental Results and Released Models}
Based on the CoT selection guidelines, for each of the 708K problems in \texttt{OmniThought}, we sample suitable CoTs to train models via SFT. To verify that our dataset is suitable for obtaining models with varying parameter sizes, we train two models initialized from the Qwen2.5 series (7B and 32B).\footnote{The two models have been released and named as DistilQwen-ThoughtX-7B and DistilQwen-ThoughtX-32B. DistilQwen-ThoughtX-7B: \url{https://huggingface.co/alibaba-pai/DistilQwen-ThoughtX-7B}
DistilQwen-ThoughtX-32B: \url{https://huggingface.co/alibaba-pai/DistilQwen-ThoughtX-32B}} Note that due to the difference in parameter size, the training sets for the two models include different CoTs based on RV and CD samplers. Detailed experimental settings are presented in Appendix~\ref{sec:exp}. We further compare the performance of our models against state-of-the-art models in the open-source community, with results summarized in Table~\ref{tab:strong}. We observe that, based on the constructed datasets and our CoT selection strategies, we are able to obtain strong LRMs that clearly outperform previous ones in the community. Furthermore, by annotating the RV and CD scores for all CoTs, our \texttt{OmniThought} dataset has the potential for obtaining strong LRMs based on LLMs of other sizes.

\subsection{Further Experimental Studies}


\begin{wraptable}{r}{0.475\textwidth}
\centering
\caption{Comparison between our trained models with different settings (averages only).}
\begin{small}
\begin{tabular}{l|ll}
\hline
\bf Setting & \bf Model (7B) & \bf Model (32B)\\
\hline
Full Dataset & 55.5 & 72.8\\
\hline
$\mu_{CD}=3$ & 44.2 & 65.1\\
$\mu_{CD}=5$ & \bf 63.4 & 74.8\\
$\mu_{CD}=7$ & 59.5 & \bf 77.5\\
$\mu_{CD}=9$ & 57.8 & 74.9\\
\hline
\end{tabular}
\end{small}
\label{tab:summary}
\end{wraptable}

We further conduct detailed studies on CoT selection strategies. For the two 7B/32B models, we set $\mu_{CD}$ to 3, 5, 7, and 9. We also utilize the entire \texttt{OmniThought} dataset as the training set. We perform SFT training on each backbone with these three datasets under identical training configurations as in the main experiments. Due to space limitations, we report the experimental results in Table~\ref{tab:more} in Appendix~\ref{sec:result} and summarize the results in Table~\ref{tab:summary}, which reports the average scores over AIME2024, MATH500, GPQA-D, and LCB V2. From the results, we observe that for the 7B model, the optimal setting is $\mu_{CD}=5$, while for the 32B model, the optimal setting is $\mu_{CD}=7$. Our findings are also consistent with the results in Table~\ref{tab:cd_exp}. We suggest that, when training a relatively small model (<10B), a modest $\mu_{CD}$ value is suitable, while for larger models, one may consider increasing $\mu_{CD}$. In addition, leveraging the entire dataset for training does not yield optimal results.

\begin{wrapfigure}{r}{0.375\textwidth}
    \centering
    \includegraphics[width=\linewidth]{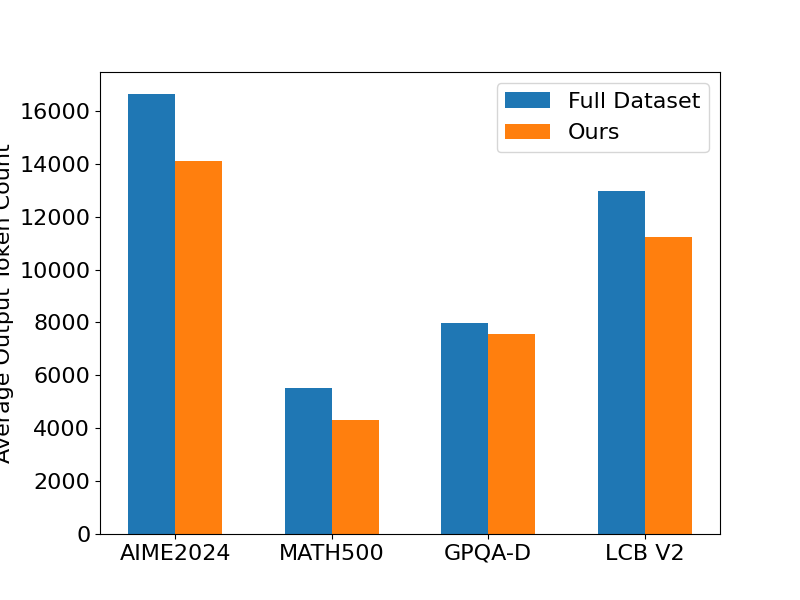}
    \caption{The average output token counts for 7B models.}
    \label{fig:tokens}
\end{wrapfigure}

To demonstrate that our approach can effectively produce more concise CoT processes, we further analyze the average output token counts for both 7B and 32B models across the four benchmarks. As shown in Figure~\ref{fig:tokens} for 7B models, our proposed method consistently reduces the output token count compared to training on the full dataset. For the 7B model, the reduction is most pronounced in the AIME2024 dataset. The reduction trend is highly similar for 32B models as well (presented in Figure~\ref{fig:tokens32b} in Appendix~\ref{sec:exp}). These results demonstrate that our approach not only maintains reasoning performance but also improves inference efficiency with shorter output tokens.

\section{Concluding Remarks}

In conclusion, the introduction of \texttt{OmniThought}, a large dataset comprising \textbf{2 million} CoT processes annotated with Reasoning Verbosity and Cognitive Difficulty scores, addresses a critical gap in the development of LRMs. Through a self-reliant curation pipeline and the demonstration of improved LRM training effectiveness using Qwen2.5 models, \texttt{OmniThought} empowers the creation of high-performing LRMs with strong reasoning abilities and optimal CoT length and difficulty levels.

\textbf{Limitations and Future Work.} Several limitations exist in this work, such as the limited scope of reasoning tasks and the static nature of the dataset construction process. These can be addressed in future work through: i) expanding the dataset to cover a broader spectrum of domains, ii) enhancing the data curation pipeline with adaptive feedback mechanisms and interactive learning loops, and iii) extending the annotations to include dynamic assessments of reasoning proficiency.

\textbf{Broader Impacts.} The broader implications of \texttt{OmniThought} extend beyond mere performance improvements. As LRMs become increasingly integrated into various applications, the construction methodology of \texttt{OmniThought} will also contribute to the development of LRMs in diverse scenarios. Nonetheless, it is essential to remain vigilant about ethical considerations, particularly regarding the use of CoTs in sensitive areas, such as the generation of reasoning processes for sensitive tasks.

\newpage
\medskip

{

}
\newpage


\appendix

\section{CoT Metadata}

\label{sec:metadata}

In the \texttt{OmniThought} dataset, each problem is associated with multiple CoTs. These CoTs are accompanied by comprehensive metadata annotations. Below is a concrete example of CoT metadata. This CoT was generated by \texttt{DeepSeek-R1} and validated as correct by \texttt{QwQ-32B}. Here, ``thought'' represents the reasoning process (CoT), ``solution'' denotes the solution derived from that process, and ``full\_response'' is the combination of both. This CoT was scored by \texttt{QwQ-32B} acting as the \texttt{Score Calculator} for Reasoning Verbosity (RV) and Cognitive Difficulty (CD), receiving an RV score of 5 and a CD score of 7.

\begin{verbatim}
{
    "thought": "TL;DR",
    "solution": "TL;DR",
    "full_response": "TL;DR",
    "teacher": DeepSeek-R1,
    "thought_correctness_verify": True,
    "Reasoning_Verbosity": {"level": 5, "judge": QwQ-32B},
    "Cognitive_Difficulty": {"level": 7, "judge": QwQ-32B}
}
\end{verbatim}

CoTs in \texttt{OmniThought} can be assembled into various dataset types according to metadata filtering requirements, thereby supporting the application of specialized training algorithms. In future work, we will continue to enrich CoT metadata, for example, by incorporating additional judge models and introducing further scoring dimensions. As the metadata becomes increasingly comprehensive, \texttt{OmniThought} will enable more diverse and customizable training regimes, facilitating model adaptation to a broader range of task scenarios.

\section{Detailed Sampling Process}

\label{sec:sample}

For simplicity, we denote:
\begin{equation}
f_1(y_{CoT})= 
\begin{cases}
\max_{i=1,\ldots,|\mathcal{D}(x)|}\bigl\lvert S^{(i)}_{CD}-\mu_{CD}\bigr\rvert, & S_{CD}\le\mu_{CD},\\[0.5em]
\max_{i=1,\ldots,|\mathcal{D}(x)|}\bigl\lvert S^{(i)}_{CD}-\mu_{CD}\bigr\rvert - \bigl(S_{CD}-\mu_{CD}\bigr), & S_{CD}>\mu_{CD},
\end{cases}
\end{equation}
\begin{equation}
f_2(y_{CoT})=\left(\max_{i=1,\ldots,\vert\mathcal{D}(x)\vert}\vert S^{(i)}_{CD}-S^{(i)}_{RV}\vert\right) - \vert S_{CD}-S_{RV}\vert.
\end{equation}

We then normalize $f_1(y_{CoT})$ and $f_2(y_{CoT})$ to obtain the corresponding probability distributions $P_1(y_{CoT})$ and $P_2(y_{CoT})$, i.e.,
\begin{equation}
P_1\bigl(y_{CoT}\bigr)=
\frac{f_1\bigl(y_{CoT}\bigr)}
{\displaystyle
\sum_{i=1}^{|\mathcal{D}(x)|}
f_1\bigl(y^{(i)}_{CoT}\bigr)
} \ \ \ \text{and}\ \ \  P_2\bigl(y_{CoT}\bigr)=
\frac{f_2\bigl(y_{CoT}\bigr)}
{\displaystyle
\sum_{i=1}^{|\mathcal{D}(x)|}
f_2\bigl(y^{(i)}_{CoT}\bigr)
}. 
\end{equation}
The final probability of selecting a CoT is defined as follows: 
\begin{equation}
\Pr(y_{CoT}) = \beta\cdot P_1\bigl(y_{CoT}\bigr) + (1-\beta)\cdot P_2\bigl(y_{CoT}\bigr)
\end{equation}
where $\beta$ is a tunable hyper-parameter that determines the respective influence of $P_1(y_{CoT})$ and $P_2(y_{CoT})$ on the final probability. In default, we set $\beta=0.5$ to strike a balance between the two factors. Given $\mu_{CD}$, we can assign selection probabilities to all CoTs corresponding to each problem in \texttt{OmniThought} and then perform sampling based on these probabilities. It is important to note that the number of CoT samples per problem can be flexibly adjusted; in other words, if multiple CoTs for a single problem are assigned high sampling probabilities, we can sample several of these high-quality CoTs simultaneously to enrich the training set.

\section{Detailed Experimental Settings}

\label{sec:exp}

\subsection{Verification on Reasoning Verbosity}
In the SFT for verification on RV, we set the global batch size to 96, the learning rate to $1 \times 10^{-5}$, and trained for 3 epochs. We utilized a single node equipped with 8 A800 GPUs (80 GB), achieving a training time of approximately 6 hours per model.

\subsection{Explorations on RQ1 and RQ2}
Similarly, in the SFT training for RQ1, we set the global batch size to 96, the learning rate to $1 \times 10^{-5}$, and trained for 3 epochs. We utilized a single node with 8 A800 GPUs (80 GB), with each model taking approximately 5 hours to train.

In the DPO training for RQ2, we set the global batch size to 96, the learning rate to $5 \times 10^{-7}$, $\beta$ to 0.1, and trained for 1 epoch. We employed a single node with 8 A800 GPUs (80 GB), and the training time was 2 hours.

\subsection{Training Strong LRMs}
For the SFT training of the 7B model, we set the global batch size to 512, the learning rate to $8 \times 10^{-5}$, and trained for 5 epochs. We utilized 8 nodes, each equipped with 8 A800 GPUs (80 GB), resulting in a training time of approximately 26 hours.

For the SFT training of the 32B model, we set the global batch size to 512, the learning rate to $3 \times 10^{-5}$, and trained for 5 epochs. We employed 8 nodes, each with 8 A800 GPUs (80 GB), achieving a training time of approximately 140 hours.

\section{Supplementary Experimental Results}

\label{sec:result}

\begin{figure}[t]
    \centering
    \includegraphics[width=.75\linewidth]{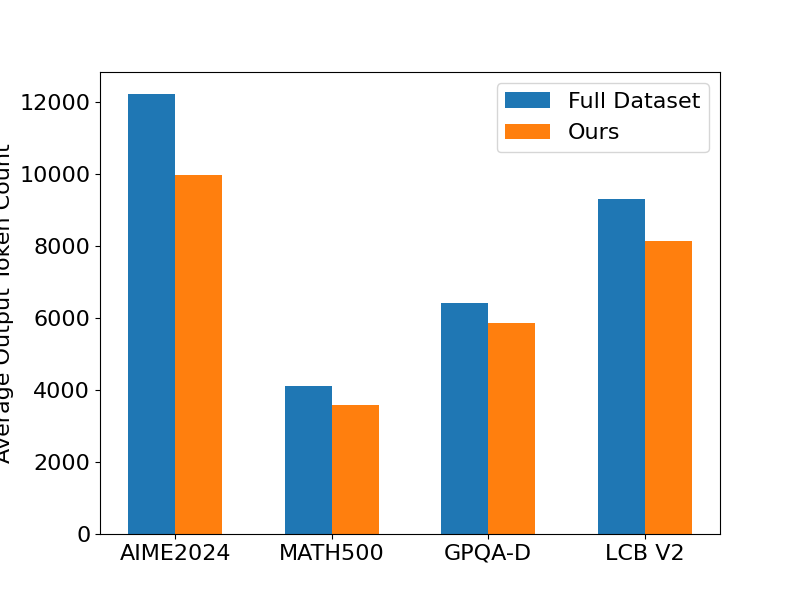}
    \caption{The average output token counts for 32B models across four testing sets.}
    \label{fig:tokens32b}
\end{figure}

Detailed experimental results on our trained models with different settings are presented in Table~\ref{tab:more}. The average output token counts for 32B models are shown in Figure~\ref{fig:tokens32b}.

\begin{table}[t]
\centering
\caption{Performance comparison between our trained models with different settings.}
\begin{small}
\begin{tabular}{l|llll|l}
\hline
\bf Setting & \bf AIME2024 & \bf MATH500 & \bf GPQA-D & \bf LCB V2 & \bf Avg.\\
\hline
\multicolumn{2}{l}{Model Size: 7B}\\
\hline
Full Dataset & 43.3 & 88.2 & 45.4 & 45.4 & 55.5\\
$\mu_{CD}=3$ & 23.3 & 79.8 & 40.4 & 33.4 & 44.2\\
$\mu_{CD}=5$ & \bf 56.7 & \bf 90.2  & \bf 50.0 & \bf 56.8 & \bf 63.4\\
$\mu_{CD}=7$  & 46.7  & 90.0  & 46.4  & 54.9 & 59.5\\
$\mu_{CD}=9$ & 43.3  & 89.6  & 45.9  & 52.6 & 57.8\\
\hline
\multicolumn{2}{l}{Model Size: 32B}\\
\hline
Full Dataset & 70.0 & 91.8 & 59.6 & 70.1 & 72.8\\
$\mu_{CD}=3$ & 56.7 & 89.6 & 54.3 & 59.8 & 65.1\\
$\mu_{CD}=5$ & 73.3 & 92.4 & 61.6 & 72.0 & 74.8\\
$\mu_{CD}=7$  & \bf 80.0 & \bf 92.6 & \bf 64.0 & \bf 73.4 & \bf 77.5\\
$\mu_{CD}=9$ & 73.3 & 92.0 & 62.6 & 71.8 & 74.9\\
\hline
\end{tabular}
\end{small}
\label{tab:more}
\end{table}

\section{Prompt Templates}

\label{sec:prompt}

The prompt templates used in this paper are listed below.

\begin{tcolorbox}[colback=gray!10!white, colframe=gray!80!black, title=Prompt Template to Validate the Correctness of CoT Processes and Solution]
You are a rigorous logical validator analyzing problem-solving components. \\
Your task is to separately assess the validity of the reasoning process and final solution. \\
Given a problem, the correct answer, a candidate reasoning process, and a candidate solution, you will:\\
\\
For SOLUTION VALIDITY: Directly comparing it to the correct answer.
\\
For REASONING PROCESS VALIDATION: \\
    a. Verify stepwise logical coherence and soundness \\
    b. Confirm all critical problem constraints are properly addressed \\
    c. Check for self-contradictions or unsupported leaps in logic \\
    d. Verify the process can actually derive the proposed solution\\
\\
Evaluation Protocol:\\
- Solution validity MUST be FALSE for any numerical mismatch or missing units \\
- Reasoning process validity requires ALL validation criteria (a-d) satisfied \\
- Both assessments must be independent: correct answer with flawed reasoning gets (False, True) \\
- Return STRICT BOOLEAN assessments for both components \\
\\

Problem: \{problem\} \\
Correct Answer: \{answer\} \\
Candidate Reasoning Process: \{reasoning process\} \\
Proposed Solution: \{solution\} \\
\\
Output Format: reasoning\_valid: bool, solution\_valid: bool 

\end{tcolorbox}

\begin{tcolorbox}[colback=gray!10!white, colframe=gray!80!black, title=Prompt Template to Calculate the RV Score]
You are an expert judge tasked with evaluating the Reasoning Verbosity of a Chain-of-Thought (CoT) for a given problem and its answer. \\
\\
Reasoning Verbosity Evaluation Focus: \\
Assess how well the CoT’s length and step complexity match the problem’s inherent difficulty. \\
An optimal chain is neither missing essential steps nor padded with needless digressions. \\
A simple question should be solved with a brief, direct chain; a challenging one may justifiably require a longer path with reflection and error-checking. \\
\\
Scoring Guidelines (0-9): \\
0-1 Minimal verbosity, straightforward expression with little to no elaboration. \\
2-3 Clear and concise reasoning with necessary explanations. \\
4-5 Moderate verbosity with detailed explanations and thorough reasoning.\\
6-7 Extensive verbosity with comprehensive justification and exploration of complex connections. \\
8-9 High verbosity with deep, exhaustive exploration of reasoning; involves extensive elaboration, nested justifications, and consideration of counterarguments or alternative perspectives. \\
\\
Given Problem, Chain-of-Thought and Answer, you will: \\
1. Analyze the Reasoning Verbosity \\
2. Determine score using the above criteria \\
3. Output ONLY the integer score (0-9) \\
\\
Problem: \{problem\} \\
Chain-of-Thought: \{thought\} \\
Answer: \{solution\} \\
\\
Final Output: Single integer between 0-9

\end{tcolorbox}

\begin{tcolorbox}[colback=gray!10!white, colframe=gray!80!black, title=Prompt Template to Calculate the CD Score]
You are an expert judge assessing the Cognitive Difficulty of a Chain-of-Thought (CoT) for a given problem and its answer. \\
\\

Cognitive Difficulty Evaluation Focus: \\
The level of reasoning competence required for a model to follow and reproduce the chain faithfully. \\
Judge the reasoning approach, techniques, and overall difficulty. \\
Higher scores correspond to more advanced concepts, abstractions, or multi-layer reasoning patterns. \\
\\
Scoring Guidelines (0-9): \\
0-1 Elementary facts or a single trivial operation. \\
2-3 Multi-step arithmetic, explicit enumeration, basic rule chaining. \\
4-5 Early-undergraduate logic/algebra; one non-obvious insight. \\
6-7 Advanced undergraduate techniques (determinants, dynamic programming, layered code reasoning, etc). \\
8-9 Graduate-level abstraction, nested proofs, intricate algorithmic analysis. \\
\\
Given Problem, Chain-of-Thought and Answer, you will: \\
1. Analyze the Cognitive Difficulty \\
2. Determine score using the above criteria \\
3. Output ONLY the integer score (0-9) \\
\\
Problem: \{problem\} \\
Chain-of-Thought: \{thought\} \\
Answer: \{solution\} \\
\\
Final Output: Single integer between 0-9
\end{tcolorbox}

\section{Examples of CoT Processes}

\label{sec:cotsample}
In this section, we present examples of RV and CD across different level intervals in Table \ref{table:rv-example} and Table \ref{table:cd-example}, respectively.
For ease of understanding, we have summarize the step-by-step procedures in CoTs of Table \ref{table:cd-example}.

\begin{table}[ht]
\centering
\caption{CoT examples for \(\displaystyle \int x e^x\,dx\) at varying RV levels.}
\begin{tabular}{|c|p{12cm}|}
\hline
RV Level & CoT \\
\hline
0–1 & The integral evaluates immediately to \((x-1)e^x + C\). \\
\hline
2–3 & Using integration by parts, let \(u=x\) and \(dv=e^x\,dx\), which yields \(du=dx\) and \(v=e^x\). Hence,  
\[
\int x e^x\,dx = x e^x - \int e^x\,dx = x e^x - e^x + C.
\] \\
\hline
4–5 & Select the integration-by-parts formula:  
\[
u = x,\quad dv = e^x\,dx
\quad\Longrightarrow\quad
du = dx,\quad v = e^x.
\]  
Then  
\[
\int x e^x\,dx = x e^x - \int e^x\,dx.
\]  
Since \(\int e^x\,dx = e^x + C\), it follows that  
\[
\int x e^x\,dx = x e^x - e^x + C = (x-1)e^x + C.
\] \\
\hline
6–7 & Recall the identity \(\int u\,dv = uv - \int v\,du\). Setting \(u=x\), \(dv=e^x\,dx\) gives \(du=dx\) and \(v=e^x\). Thus,  
\[
\int x e^x\,dx = x e^x - \int e^x\,dx = x e^x - e^x + C_1 = (x-1)e^x + C.
\]  
Verification by differentiation:  
\[
\frac{d}{dx}\bigl[(x-1)e^x\bigr] = e^x + (x-1)e^x = x e^x.
\] \\
\hline
8–9 & Method 1 (integration by parts): let \(u=x\), \(dv=e^x\,dx\); then  
\[
\int x e^x\,dx = x e^x - \int e^x\,dx = (x-1)e^x + C.
\]  
Method 2 (series expansion): express  
\[
e^x = \sum_{n=0}^\infty \frac{x^n}{n!}
\;\Longrightarrow\;
x e^x = \sum_{m=1}^\infty \frac{x^m}{(m-1)!}.
\]  
Integrating termwise yields  
\[
\int x e^x\,dx = \sum_{m=1}^\infty \frac{x^{m+1}}{(m-1)!\,(m+1)} = (x-1)e^x + C.
\]  
Generalization: \(\displaystyle \int x^n e^x\,dx = x^n e^x - n\int x^{n-1}e^x\,dx.\) \\
\hline
\end{tabular}
\label{table:rv-example}
\end{table}

\begin{table}[ht]
\centering
\caption{CoT examples at varying CD levels.}
\begin{tabular}{|p{2cm}|p{3cm}|p{6cm}|p{2cm}|}
\hline
CD level & Problem & CoT (Summarized) & Answer \\ \hline
0–1 & Compute \(2 + 3\). &
1. Recognize that the operation is a simple sum of two integers.\par
2. Observe that \(2 + 3 = 5\). & \(5\) \\ \hline
2–3 & Compute the sum of the first five multiples of \(7\). &
1. List the first five multiples of \(7\): \(7, 14, 21, 28, 35\).\par
2. Compute their sum: \(7 + 14 + 21 + 28 + 35 = 105\). & \(105\) \\ \hline
4–5 & Simplify \(\frac{x^2 - 4}{x - 2}\) and evaluate at \(x = 5\). &
1. Note that \(x^2 - 4 = (x - 2)(x + 2)\).\par
2. Cancel the common factor \((x - 2)\), obtaining \(x + 2\).\par
3. Substitute \(x = 5\): \(5 + 2 = 7\). & \(7\) \\ \hline
6–7 & Determine the time complexity of \(T(n) = 2\,T(n/2) + n\). &
1. Identify \(a=2\), \(b=2\), \(f(n)=n\).\par
2. Compute \(n^{\log_2 2} = n\).\par
3. Since \(f(n)=\Theta(n)\), this is Case 2 of the Master theorem.\par
4. Conclude \(T(n)=\Theta(n\log n)\). & \(\Theta(n\log n)\) \\ \hline
8–9 & Prove that there are infinitely many prime numbers. &
1. Assume, for contradiction, primes are finite \(\{p_1,\dots,p_n\}\).\par
2. Construct \(N = p_1 p_2 \cdots p_n + 1\).\par
3. \(N\) leaves remainder 1 mod any \(p_i\).\par
4. By the Fundamental Theorem of Arithmetic, \(N\) has a prime divisor not in the list.\par
5. Contradiction implies infinitely many primes. & Infinitely many primes \\ \hline
\end{tabular}
\label{table:cd-example}
\end{table}

\end{document}